\documentclass[sigconf,natbib=true]{acmart}
\AtBeginDocument{%
  }

\setcopyright{acmlicensed}
\copyrightyear{2018}
\acmYear{2018}
\acmDOI{XXXXXXX.XXXXXXX}
\acmConference[Conference acronym 'XX]{Make sure to enter the correct
  conference title from your rights confirmation email}{June 03--05,
  2018}{Woodstock, NY}
\acmISBN{978-1-4503-XXXX-X/2018/06}
\usepackage{color}
\usepackage{adjustbox}
\usepackage{rotating}
\usepackage{makecell}
\usepackage[table]{xcolor}
\usepackage{xspace}
\usepackage{multirow,multicol,booktabs,verbatim,wrapfig}
\usepackage{float}
\usepackage{tabularx}
\usepackage{xcolor}
\usepackage{multicol}
\usepackage{subcaption}
\usepackage{caption}
\usepackage{xspace}

\definecolor{em}{RGB}{215,229,242}
\definecolor{gree}{RGB}{232,245,233}
\definecolor{yell}{RGB}{254,246,219}
\definecolor{redd}{RGB}{254,229,219}
\newcommand{\cem}{\cellcolor{em}}
\newcommand{\gre}{\cellcolor{gree}}
\newcommand{\yel}{\cellcolor{yell}}
\newcommand{\reddd}{\cellcolor{redd}}

\newcommand{\bs}[1]{{\textbf{#1}}}

\usepackage{algorithm}
\usepackage{algpseudocode}
\usepackage{pgfplots}
\pgfplotsset{compat=1.18}

\newcommand{\ours}{\textbf{AMA}\xspace}

\begin{document}

\title{Enhancing Conversational Agents via Task-Oriented \\Adversarial Memory Adaptation}

\author{
    Yimin Deng\textsuperscript{1,2}, 
    Yuqing Fu\textsuperscript{2}, 
    Derong Xu\textsuperscript{2,3},
    Yejing Wang\textsuperscript{2},
    Wei Ni\textsuperscript{2,4}, 
    Jingtong Gao\textsuperscript{2},
     Xiaopeng Li\textsuperscript{2}\\
     Chengxu Liu\textsuperscript{1},
     Xiao Han\textsuperscript{5},
     Guoshuai Zhao\textsuperscript{1}$^{\dagger}$,
     Xiangyu Zhao\textsuperscript{2}$^{\dagger}$,
     Li Zhu\textsuperscript{1}$^{\dagger}$,
      Xueming Qian\textsuperscript{1}
    \\
    \textsuperscript{1}Xi'an Jiaotong University, \textsuperscript{2}City University of Hong Kong\\
    \textsuperscript{3}University of Science and
Technology of China,
    \textsuperscript{4}Zhejiang University\\
   %
    \textsuperscript{5}Zhejiang University of Technology 
   \\
  \small{
  \texttt{
  \href{mailto:dymanne@stu.xjtu.edu.cn}{dymanne@stu.xjtu.edu.cn},
    \href{mailto:guoshuai.zhao@xjtu.edu.cn}{guoshuai.zhao@xjtu.edu.cn},
     \href{mailto:zhuli@xjtu.edu.cn}{zhuli@xjtu.edu.cn}}},
    \href{mailto:xianzhao@cityu.edu.hk}{xianzhao@cityu.edu.hk}
  }
  \thanks{\textsuperscript{\dag}Corresponding author.}
   
\renewcommand{\shortauthors}{Trovato et al.}

\begin{abstract}

  Conversational agents struggle to handle long conversations due to context window limitations. Therefore, memory systems are developed to leverage essential historical information. Existing memory systems typically follow a pipeline of offline memory construction and update, and online retrieval. 
   Despite the flexible online phase, the offline phase remains fixed and task-independent. 
  In this phase, memory construction operates under a predefined workflow and fails to emphasize task-relevant information. Meanwhile, memory updates are guided by generic metrics rather than task-specific supervision. This leads to a misalignment between offline memory preparation and task requirements, which undermines downstream task performance. To this end, we propose an \textbf{A}dversarial \textbf{M}emory \textbf{A}daptation mechanism~(\textbf{AMA}) that aligns memory construction and update with task objectives by simulating task execution. Specifically, first, a challenger agent generates question–answer pairs based on the original dialogues. The constructed memory is then used to answer these questions, simulating downstream inference. Subsequently, an evaluator agent assesses the responses and performs error analysis. Finally, an adapter agent analyzes the error cases and performs dual-level updates on both the construction strategy and the content. Through this process, the memory system receives task-aware supervision signals in advance during the offline phase, enhancing its adaptability to downstream tasks. \textbf{AMA} can be integrated into various existing memory systems, and extensive experiments on long-dialogue benchmark LoCoMo demonstrate its effectiveness.
\end{abstract}



\received{20 February 2007}
\received[revised]{12 March 2009}
\received[accepted]{5 June 2009}

\maketitle

\section{Introduction}
Conversational agents are essential for delivering personalized and efficient digital services~\cite{huq2024dialogue,deng2023survey,ni2023recent,sigir25from,sigir25enhancing}. With the advancement of artificial intelligence and the increasing complexity of user intents, they are frequently required to handle long conversations~\cite{algherairy2024review,25sigirseeking,sigir25search}. In such scenarios, tracking user preferences and maintaining knowledge consistency becomes challenging due to the limited context window of large language models~(LLMs). In response to this issue, memory systems have been developed to support downstream tasks~\cite{hu2026memoryageaiagents,pan2025memory,zhong2024memorybank}.
\begin{figure}[t]
    \centering
    \centering
   \includegraphics[height=0.4\textwidth]{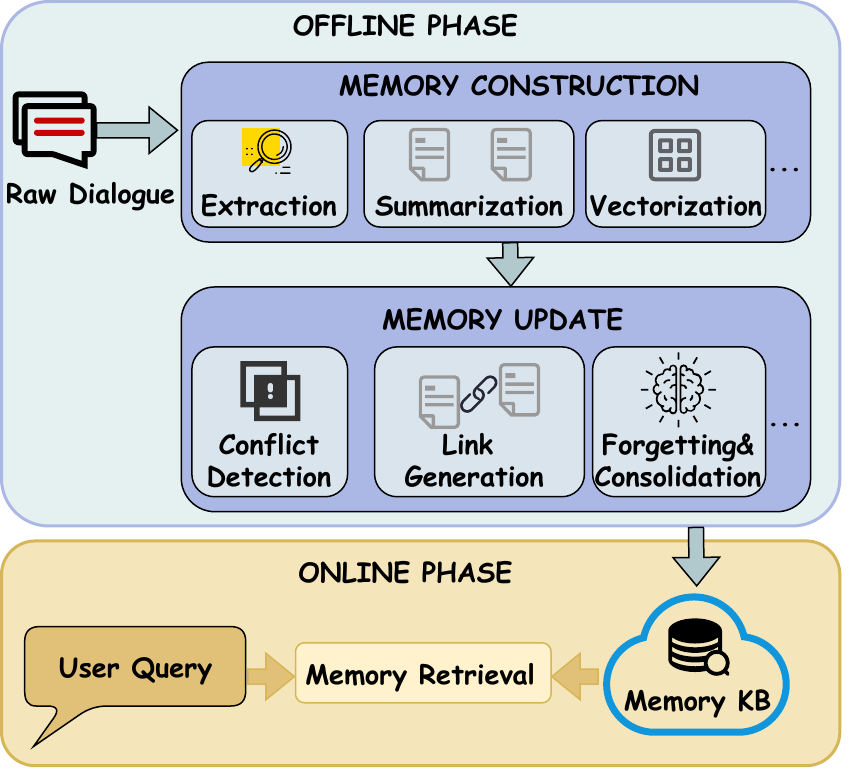}
    \caption{A typical memory system pipeline.
    }
    \label{setting}
\end{figure}

Memory systems are intended to capture key information from raw dialogue and support task-specific retrieval, playing a crucial role in preserving the consistency and stability of multi-turn conversations~\cite{wu2025human,zhang2025survey}. As shown in Figure~\ref{setting}, existing memory systems typically follow a pipeline involving \textbf{offline} memory construction and update, followed by \textbf{online} memory retrieval~\cite{du2025rethinking,hu2026memoryageaiagents,xu2025mem}. In the offline phase, the system first organizes dialogue content into structured entries to facilitate storage in the memory repository. As it constructs new memory entries, the system dynamically updates existing memory by deleting outdated content, modifying existing entries, or adding new links. In the online phase, it retrieves relevant memory entries for specific tasks. While online retrieval is flexibly tailored to downstream requirements, the offline memory preparation process is not explicitly task-driven. This leads to a misalignment between the two phases, which limits the system’s ability to generalize in real-world scenarios.

Specifically, existing approaches have developed a series of strategies for memory construction. They adopt text chunking, temporal knowledge graphs (TKGs), or vector databases to store dialogue information in the form of entries or units~\cite{rasmussen2025zep,zhong2024memorybank,packer2023memgpt,fang2025lightmem}, enabling lightweight and efficient memory storage. When processing raw dialogues, they employ techniques such as entity extraction, text summarization, and text vectorization to identify and retain key information~\cite{wang2025recursively,xu2025mem,fang2025lightmem,nan2025nemori}.
However, these strategies are manually designed and remain fixed throughout the construction process, lacking the ability to adapt to specific requirements.

In parallel, approaches focusing on memory update have also made significant progress. These approaches are designed to keep memory accurate and up-to-date. One common technique involves conflict detection, where the system identifies inconsistencies between new input and existing memory entries~\cite{xu2025mem,hu2026memoryageaiagents}. Upon detecting a conflict, it resolves it by analyzing differences and considering more recent or trustworthy information, thereby preserving memory consistency~\cite{fang2025lightmem,pan2025memory}. Another type of strategy involves new link generation, which enables the system to create links between memory units based on semantic similarity or contextual relevance. It allows the system to better capture the associations among events or entities~\cite{xu2025mem,rasmussen2025zep}. In addition, some approaches are inspired by human memory mechanisms~\cite{jimenez2024hipporag,zhong2024memorybank,lee2024human}, incorporating processes such as reinforcement of essential or frequently accessed information and forgetting of outdated or low-utility content. These biologically inspired operations help the system maintain a compact memory over time. 
While these techniques attempt to improve memory quality, they rely on generic metrics and lack task-specific supervision signals. Consequently, the update processes are not directly aligned with the target task.

However, preparing memory for specific task scenarios is crucial.
On the one hand, since tasks differ in information preference, general memory construction may result in misaligned extraction. For example, temporal reasoning requires the awareness of time-related information, whereas multi-hop reasoning demands the ability to establish connections across events~\cite{2025para,xiong2024large}.
On the other hand, shifting memory updates toward a task-oriented approach is necessary, as it can directly enhance the quality of memory for the target task.
This is similar to how humans reinforce knowledge by solving domain-specific exercises, which helps to consolidate relevant information more effectively.
Despite the importance of task-oriented memory adaptation, accessing task supervision remains challenging, as the memory system lacks direct interaction with the task during the offline phase.
This highlights the motivation for developing techniques for simulating task feedback.
Moreover, even when such feedback is available, it is often fragmented and expressed in unstructured forms (e.g., natural language), making it necessary to develop appropriate strategies to update the memory system in alignment with task requirements.

To this end, we propose an \textbf{A}dversarial \textbf{M}emory \textbf{A}daptation mechanism~(\textbf{AMA}) that guides the memory system to coordinate both its construction and update process with task requirements during memory preparation.
Inspired by adversarial principles, the proposed mechanism guides memory adaptation by simulating the task execution process and establishing a feedback loop for iterative refinement, while being seamlessly integrable with existing memory systems.
Specifically, during memory construction, a challenger agent first generates question–answer pairs from the original dialogue in a task-specific manner. The constructed memory is then used to answer these questions. Subsequently, an evaluator agent assesses the system’s performance by scoring the answers and identifying failure cases.
Finally, an adapter agent performs the dual-level memory adaptation. It (i) updates the existing memory content to incorporate missing information and (ii) adjusts the memory construction strategy to better align with the task's demands.
In summary, we simulate task demands and incorporate in-process evaluation during offline memory construction to promote alignment with online retrieval.
To ensure that the memory system evolves in a task-aligned manner, we simulate task execution through QA-based interaction, allowing the memory extraction process to be guided by what knowledge the task actually requires.

To summarize, our contributions are as follows:

\begin{list}{$\bullet$}{
  \leftmargin=1em
  \itemsep=0pt
  \topsep=0pt
  \parsep=0pt
  \partopsep=0pt
}
  \item We identify the importance of task-specific memory adaptation and propose an \textbf{A}dversarial \textbf{M}emory \textbf{A}daptation~(\ours) mechanism that enhances the generalizability of memory systems and can be seamlessly integrated into existing memory systems.
  \item We propose a dual-level update strategy that jointly optimizes memory content and construction strategy to ensure consistency between offline memory preparation and online retrieval.
  \item Extensive experiments on LoCoMo with three memory systems and two backbone models validate the effectiveness of \ours.
\end{list}
 \begin{figure*}[t]
    \centering
    \resizebox{1\linewidth}{!}{
    \includegraphics[height=0.7\textwidth]{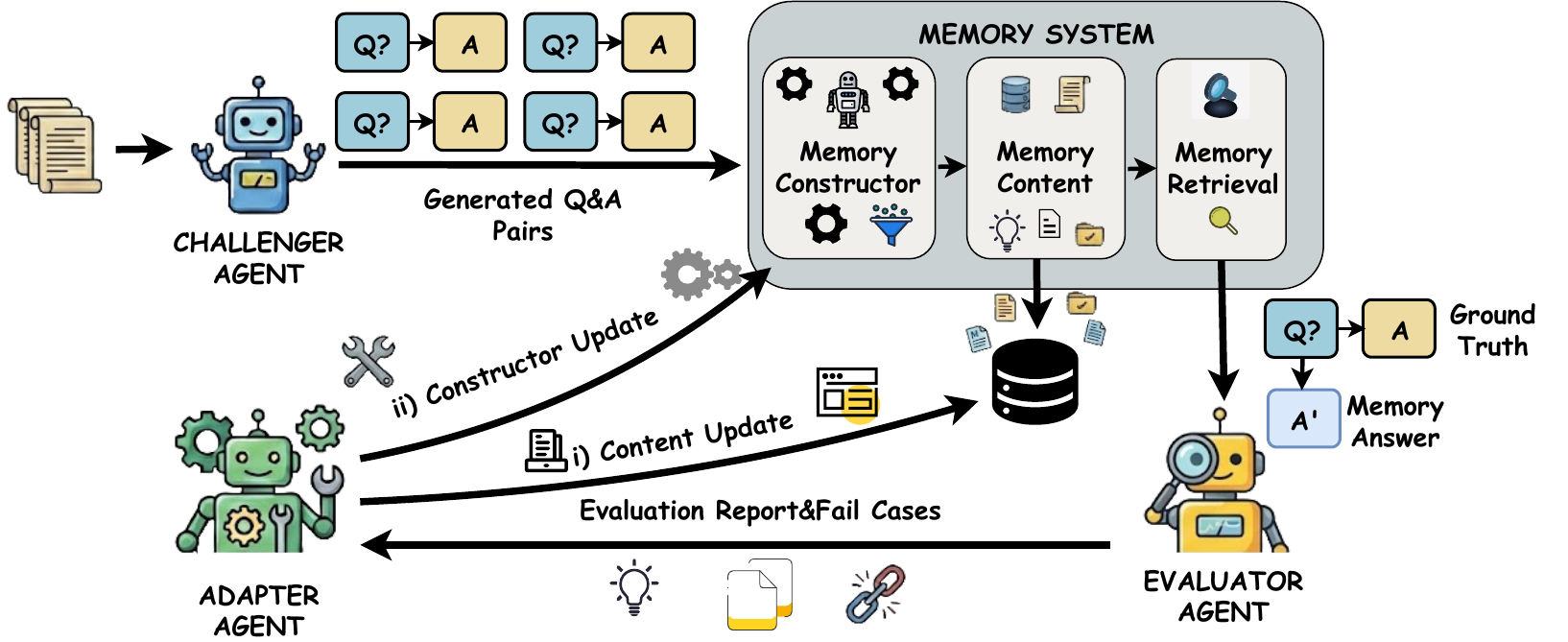}}
    \caption{The overall architecture of our model. 
    }
    \label{model}
    \vspace{-3mm}
\end{figure*}
\section{Preliminaries}
In this section, we provide an overview of memory system architecture, which serves as the necessary context for our proposed approach.
A typical memory system is composed of three key modules: memory construction, memory update, and memory retrieval~\cite{xiong2025enhancing,zhang2025ask}. These modules work sequentially and collaboratively to support the storage, management, and utilization of key facts, thereby enhancing the contextual understanding of long, multi-turn conversations for conversational agents~\cite{lin2025hippomm,liu2025agentcf++,song2024moviechat,sigir25unveil,wang2025more}.

\noindent\textbf{Memory Construction.} The memory construction module is designed to organize raw multi-turn dialogues into structured memory entries, thereby supporting online retrieval. It takes the original dialogue $D$ as input and transforms it into a set of memory entries $M = \{m_1, m_2, \dots\}$ through a series of preprocessing steps. 

This process typically involves: 
\begin{list}{$\bullet$}{
  \leftmargin=1em
  \itemsep=0pt
  \topsep=0pt
  \parsep=0pt
  \partopsep=0pt
}
  \item \textbf{Entity and event extraction}: The system captures key entities (e.g., persons, locations, objects) and events from the dialogue for knowledge grounding and information compression. This step is usually performed using entity extraction tools or LLMs. 
  \item \textbf{Semantic compression and summarization}: It condenses long or repetitive utterances into concise representations, preserving key intent and factual content while reducing noise. This process is typically performed using a language model. 


  \item \textbf{Vector encoding}: It transforms the processed memory content into dense semantic vectors using pretrained embedding models, enabling effective retrieval and alignment with queries.

\end{list}
Each memory entry is organized with multiple structured attributes such as timestamps, keywords, and original context segments, enabling efficient indexing and retrieval within the memory system.

\noindent\textbf{Memory Update.} After creating a sequence of memory entries ${m_1, …, m_i}$, when a new memory entry $m_{i+1}$ is constructed, the module selectively modifies previous entries based on relevance or semantic similarity. It performs conflict detection to identify redundant or contradictory entries,  removes outdated or irrelevant information, and eliminates repetitive content that may lead to knowledge inflation. These operations ensure that the memory repository remains compact, consistent, and semantically relevant, thereby preserving the quality and efficiency of downstream tasks.

\noindent\textbf{Memory Retrieval.} In this stage, the structured memory entries are utilized in downstream tasks such as question answering~\cite{li2025graph}. The retrieval process employs semantic similarity to identify the most relevant information from the memory repository with respect to the specific task, thereby enabling the model to generate responses with high-quality contextual understanding~\cite{wang2025mem}.

Through the coordinated operation of the three modules, the memory system is able to maintain consistent understanding of historical dialogue, perform dynamic updates, and support efficient access, even in long, multi-turn conversations. 
\section{Method}

In this section, we first present the overall framework and then provide a detailed description of each component of our \ours. We then conclude by illustrating the complete workflow, demonstrating how they are seamlessly integrated into the existing memory system to enhance and extend its capabilities.

\subsection{Overall Framework}
In this section, we present the overall architecture of our model. As illustrated in Figure~\ref{model}, the model consists of three main components. First, a challenger agent generates question–answer (QA) pairs based on the original dialogue. These questions are then fed into the memory system to generate memory-grounded answers. Subsequently, the QA pairs, along with the system's responses, are passed to an evaluator agent, which performs assessment and error analysis. Finally, an adapter agent produces a dual-level update strategy based on the error record. The existing memory content and the next-step extraction strategy will be jointly updated.
\subsection{QA Generator}


Existing memory systems suffer from a lack of task supervision signals and fail to effectively address task requirements~\cite{zhang2025survey,hu2026memoryageaiagents}. To better align with the requirements of real-world application scenarios, leveraging task feedback to guide model behavior is beneficial. However, the feedback is only available at the online retrieval stage and is difficult to obtain during construction. To this end, we introduce a \textbf{Challenger} agent to simulate task execution and identify potential failure patterns in advance. It automatically constructs question-answer pairs based on dialogue content. The input to this agent is the original dialogue text used for memory construction. Its core functionality lies in extracting key facts from the dialogue and generating corresponding questions and accurate answers around those facts.

Specifically, we use the conversation session as the basic unit of context. Combined with a carefully designed instruction (denoted as instruction $I_q$), we prompt a LLM to generate question-answer pairs that are closely aligned with the given context. Each QA pair focuses on important facts within the dialogue, ensuring that the questions are representative and the answers are accurate. Given a dialogue $D$ and an instruction prompt $I_q$ \footnote{Due to space constraints, all prompts used in our method will be provided in the anonymized code repository.}, the QA generator leverages a LLM agent which serves as a $\text{Challenger}$ to produce a set of question-answer pairs $(q, a)$ as follows:
\begin{equation}\label{eq1}
\{(q_1, a_1), (q_2, a_2), \dots, (q_k, a_k)\} = \text{Challenger}(I_q, D)    
\end{equation}

where $q$ denotes a question that targets a key fact in the dialogue, and $a$ is the corresponding correct answer. 

These automatically generated questions not only capture the key information points within the conversation, but also serve as a valuable reference for evaluating the quality of generated memory. By comparing the model-generated memory with these reference QA pairs, we can more objectively assess the memory system’s ability to capture and represent essential information.
\subsection{Memory Evaluator}


Since memory quality has a significant impact on downstream tasks, continuous evaluation of memory quality during construction is essential. Existing approaches typically rely on generic metrics such as information redundancy~\cite{xu2025mem,fang2025lightmem}, which struggle to provide evaluations aligned with task requirements. Given the questions generated in the previous step, the accuracy of question answering serves as an effective metric for memory quality, particularly when the memory is applied to downstream reasoning tasks. To evaluate the effectiveness of the constructed memory under specific task requirements, we design and implement a memory \textbf{Evaluator} module. This component aims to generate an answer $\hat{a}$ based on the constructed memory representation $m$ and a given question $q$, and then compare it with the reference answer $a$ to determine whether the memory supports correct reasoning.

Specifically, given a reasoning instruction $I_r$, a constructed memory representation $M$ and a task-specific question $q$, the evaluator first generates an answer $\hat{a}$:
\begin{equation}\label{eq2}
\hat{a} = \text{Evaluator}(I_r,M, q)    
\end{equation}

 Then it assesses the consistency between $\hat{a}$ and the ground-truth answer $a$, returning a Boolean value $r \in \{0, 1\}$ that indicates whether the memory meets the task requirements. 
In addition, the evaluator outputs a defect description $\delta$ with analysis instruction $I_a$, which identifies potential issues in the memory, such as missing information, factual inaccuracies, or ambiguous expressions.
\begin{equation}\label{eq3}
   r,\delta = \text{Evaluator}(I_a, q, \hat{a}, a) 
\end{equation}

Through this mechanism, the system performs quantitative evaluation of memory quality and provides targeted feedback that serves as guidance for subsequent memory refinement and updates.

\subsection{Memory Adapter}
After obtaining task feedback, it is crucial to devise appropriate strategies to adapt the system based on this feedback. However, the acquired feedback is often fragmented and text-based, making it difficult to directly use as a supervision signal. To ensure continual improvement of the memory system, we introduce an \textbf{Adapter} module that performs task-oriented updates based on error records identified during evaluation. For questions that are difficult to answer based on the existing memory, the adapter generates update strategies according to the type of error, with the goal of improving memory quality. This step enables the system to revise incorrect or missing knowledge and adapt its strategy for subsequent information extraction.

The adapter applies a dual-level update strategy: (i) updating the extraction strategy $\pi$, and (ii) updating the memory content $M$.

\paragraph{i. Updating the Extraction Strategy.}  
Given the observed errors, the adapter refines the extraction strategy $\pi$ to improve the relevance and quality of retrieved content. This is achieved by either modifying the prompt used by the constructor module or by pre-filtering the input content before extraction. 
Based on the current extraction strategy $\pi$, an instruction $I_\pi$ is first constructed to guide the extraction behavior. Given the instruction $ I_\pi$ and the error record $E={r,\delta}$, the adapter formulates a prompt and employs LLMs to generate an optimized strategy adjustment.

Formally, the adjustment $\Delta \pi$ is obtained as:
\begin{equation}\label{eq4}
\Delta \pi = \text{Adapter}(I_\pi, E)
\end{equation}

The updated extraction strategy is then generated by adding the adjustment to the original strategy:
\begin{equation}\label{eq5}
\pi' = \pi + \Delta \pi
\end{equation}
this process allows the system to iteratively refine the extraction policy in response to observed failures, improving its ability to provide relevant content in future interactions. According to the system's extraction strategy, this update mechanism can be flexibly integrated. Specifically, for prompt-based extraction implemented by LLMs, the generated improvement instructions in $\pi$ can be used to optimize the prompts. For memory systems that operate on raw text, effective information can be filtered through question answering to form high-quality summaries. In addition, the accuracy can serve as a criterion for activating memory reconstruction.

\paragraph{ii. Updating the Memory Content.}  
To address factual incompleteness or errors detected in the memory, the adapter adds supplemental knowledge extracted from the error record. Specifically, if an answer is incorrect due to missing or outdated information, the adapter constructs an instruction $I_m$ to prompt LLMs for factual correction. This instruction involves guidance for summarizing the missing knowledge based on the observed error.

Given the instruction $I_m$ and the error record $E=\{r,\delta\}$, the LLM generates a factual supplement:
\begin{equation}\label{eq6}
\Delta M = \text{Adapter}(I_m, E)
\end{equation}
the corresponding factual correction is appended to the memory. The updated memory is defined as:
\begin{equation}\label{eq7}
  M' = M + \Delta M  
\end{equation}

where $\Delta M$ consists of new facts inferred from the evaluator's error analysis and serves to complete the memory, focusing on the task-relevant knowledge in failed QA cases. This module is able to identify missing information in the memory and inject it in a targeted manner, mitigating similar errors in future interactions.

 Through this dual-level update mechanism, the adapter ensures that the memory system evolves to better reflect the task distribution and improves its ability to support future question answering. The update process continues throughout memory construction, guiding the memory system to evolve toward task requirements and continuously improving the quality of the constructed memory.

\subsection{The Adversarial Memory Adaptation Algorithm}
The overall process of \ours is outlined in Algorithm~\ref{alg:memory-update}. The algorithm takes a dialogue $D$ , a memory content $M$ , and an extraction strategy $\pi$ as input. First, the challenger agent generates QA pairs $(Q, A)$ from the original dialogue $D$ (line~4). Each question $q_i$ is then submitted to the memory system, which attempts to answer it using the existing memory $M$ (lines~6-8). 
Next, the evaluator agent compares the response $\hat{A}$ with the ground-truth answers $A$ and produces an error record $E$ (line~10). Based on the error record, the adapter agent generates a dual-level update strategy for modifying the memory content $\Delta M$ and adjusting the extraction strategy $\Delta \pi$ (line~12). 
Finally, the system applies the updates to obtain an improved memory $M'$ and a revised extraction strategy $\pi'$ (lines~14-15), which are returned for subsequent memory construction (line~16).

\begin{algorithm}
\caption{The Adversarial Memory Adaptation Mechanism}
\label{alg:memory-update}
\begin{algorithmic}[1]
\State \textbf{Input:} Dialogue $D$, current memory $M$, extraction strategy $\pi$
\State \textbf{Output:} Updated memory $M'$, updated extraction strategy $\pi'$

\Statex
\State // Step 1: Generate QA pairs
\State $(Q, A) \gets \text{Challenger}(D)$ (EQ.~\ref{eq1})

\Statex
\State // Step 2: Query memory system
\For{each $q_i \in Q$}
    \State $\hat{a}_i \gets \text{Evaluator}(M, q_i)$ (EQ.~\ref{eq2})
\EndFor

\Statex
\State // Step 3: Evaluate answers
\State $E \gets \text{Evaluator}(Q, A, \hat{A})$ (EQ.~\ref{eq3})

\Statex
\State // Step 4: Generate dual update strategy
\State $(\Delta M, \Delta \pi) \gets \text{Adapter}(E)$ (EQ.~\ref{eq4},~\ref{eq6})

\Statex
\State // Step 5: Apply updates
\State $\pi' \gets \pi + \Delta \pi$ (EQ.~\ref{eq5})
\State $M' \gets M + \Delta M$ (EQ.~\ref{eq7})

\Statex
\State \Return $M'$, $\pi'$
\end{algorithmic}
\end{algorithm}

\section{Experiments}

In this section, we conduct a series of experiments to investigate the following research questions~(RQs):
\begin{list}{$\bullet$}{
  \leftmargin=1em
  \itemsep=0pt
  \topsep=0pt
  \parsep=0pt
  \partopsep=0pt
}
\item \textbf{RQ1:} How does our proposed \ours mechanism jointly improve the memory quality and the memory system's adaptability?
\item \textbf{RQ2:} How can our mechanism generalize across different memory systems and backbone models?
\item \textbf{RQ3:} What are the effects of each component of \ours on the memory system? How do content-level and strategy-level updates influence memory quality respectively?
\item \textbf{RQ4:} How does the memory system evolve with \ours?
\item \textbf{RQ5:} How does the number of QA pairs generated by the challenger agent affect the final performance?
\item \textbf{RQ6:} Does \ours improve the quality of final answers through adversarial strategies?
\end{list}
\subsection{Experimental Settings}
 \textbf{Dataset}.
 We conduct experiments on the long-context conversational benchmark LoCoMo~\cite{maharana2024evaluating} to systematically evaluate the quality of the memory system. Locomo consists of 10 dialogues with an average of over 580 turns and more than 16K tokens, and includes various categories of questions~(single-hop, multi-hop, temporal reasoning, and open-domain knowledge) to comprehensively assess memory quality for long-term dialogue. 

\begin{table}[tb!]
\centering
\caption{
    Statistics of LOCOMO Dataset.
}
\label{tab:stat_locomo}
\vspace{-5pt}

\begin{tabular}{c|c|c|c|c}
\hline
\textbf{Type} & \textbf{Multi Hop} & \textbf{Temporal} & \textbf{Open} & \textbf{Single Hop}\\
\hline

\textbf{Counts} &282&321&96&841 \\

\hline
\end{tabular}
\end{table}

\noindent\textbf{Baselines}. For comparative evaluation, we select the latest and representative agentic methods as the baselines. 
\begin{list}{$\bullet$}{
  \leftmargin=1em
  \itemsep=0pt
  \topsep=0pt
  \parsep=0pt
  \partopsep=0pt
}
\item ReadAgent~\cite{lee2024human} is an LLM-based agent system inspired by human reading strategies, following a three-stage pipeline consisting of segmentation, summarization, and retrieval. 
\item MemoryBank~\cite{zhong2024memorybank} proposes an anthropomorphic long-term memory mechanism. It operates by constructing a comprehensive memory repository that stores chronological conversation records, hierarchical event summaries, and dynamic user portraits.
\item MemGPT~\cite{packer2023memgpt} is inspired by the concept of virtual memory management in operating systems, and proposes a virtual context management mechanism for LLMs.

\end{list}

To demonstrate the compatibility of our proposed AMA mechanism, we incorporate it into three state-of-the-art baselines and conduct corresponding comparative experiments.
\begin{list}{$\bullet$}{
  \leftmargin=1em
  \itemsep=0pt
  \topsep=0pt
  \parsep=0pt
  \partopsep=0pt
}
\item A-MEM~\cite{xu2025mem} is a memory system that constructs structured notes to integrate multi-dimensional semantic attributes, including contextual descriptions, keywords, and tags.
\item LightMem~\cite{fang2025lightmem} combines the three-level human memory model with lightweight design, forming a structure of Perceptual, Short-Term, and Long-Term Memory.
\item Nemori~\cite{nan2025nemori} segments raw dialogues into semantically coherent chunks, which are then transformed into structured situational memories containing both concise titles and detailed narratives.
\end{list}

\begin{table*}[tb!]
\centering
\caption{
    Experimental results on the LoCoMo dataset, evaluating four types of QA tasks. Results are reported using F1 and BLEU-1 (\%) scores. Our method \ours (highlighted in blue) is tested with three memory systems and two LLM backbones.  The best scores are marked in bold. * denotes the average over three random runs and the improvements compared to the corresponding baseline are statistically significant ($p <
0.05$ under a t-test). $^{\ddag}$ marks results based on three dialogues due to computational cost of the baseline on GPT-4o. $^{\dag}$ indicates results obtained from the original paper of A-MEM~\cite{xu2025mem}. 
}
\label{tab:main}
\vspace{-5pt}
\resizebox{\textwidth}{!}{%
\begin{tabular}{c|l|cc|cc|cc|cc|cc}
\hline
\multirow{2}{*}{\textbf{Model}} & \multirow{2}{*}{\textbf{Method}} 
& \multicolumn{2}{c|}{\textbf{Multi Hop}} 
& \multicolumn{2}{c|}{\textbf{Temporal}} 
& \multicolumn{2}{c|}{\textbf{Open Domain}} 
& \multicolumn{2}{c|}{\textbf{Single Hop}} 
& \multicolumn{2}{c}{\textbf{Average}} \\ \cline{3-12}
& & \textbf{F1} & \textbf{BLEU} & \textbf{F1} & \textbf{BLEU} & \textbf{F1} & \textbf{BLEU} & \textbf{F1} & \textbf{BLEU} & \textbf{F1} & \textbf{BLEU} \\ \hline

\multirow{10}{*}{\textbf{GPT-4o-mini}} 
& \textsc{LoCoMo~\cite{maharana2024evaluating}}$^{\dag}$ & 25.02 & 19.75 & 18.41 & 14.77 & 12.04 & 11.16 & 40.36 & 29.05 &31.21 &23.26 \\
& \textsc{ReadAgent~\cite{lee2024human}}$^{\dag}$ & 9.15 & 6.48 & 12.60 & 8.87 & 5.31 & 5.12 & 9.67 & 7.66 &9.91 &7.54 \\
& \textsc{MemoryBank~\cite{zhong2024memorybank}}$^{\dag}$ & 5.00 & 4.77 & 9.68 & 6.99 & 5.56 & 5.94 & 6.61 & 5.16 &6.89 & 5.52\\
& \textsc{MemGPT~\cite{packer2023memgpt}}$^{\dag}$ & 26.65 & 17.72 & 25.52 & 19.44 & 9.15 & 7.44 & 41.04 & 34.34 &33.18 & 26.51\\\cline{2-12}
& A-MEM~\cite{xu2025mem} & 20.1 & 15.0 & 33.1 & 29.3 & 10.3 & 9.3 & 29.2 & 24.3 & 28.0 & 24.4 \\
& \cem{A-MEM+\ours}* & \cem{\textbf{21.5}} & \cem{\textbf{15.2}} & \cem{\textbf{34.7}} & \cem{\textbf{31}} & \cem{8.7} & \cem{8.3} & \cem{\textbf{35.1}} & \cem{\textbf{29.1}} & \cem{\textbf{31.5↑}} & \cem{\textbf{27.3↑}} \\
& LightMEM~\cite{fang2025lightmem} & 37.41 & 25.90 & 54.68 & 40.77 & 22.83 & 16.96 & 49.93 & 39.17 & 46.94 & 35.69 \\
& \cem{LightMEM+\ours}* & \cem{37.17} & \cem{25.74} & \cem{\textbf{55.97}} & \cem{\textbf{42.57}} & \cem{\textbf{24.22}} & \cem{\textbf{18.17}} & \cem{49.88} & \cem{\textbf{39.84}} & \cem{\textbf{47.22↑}} & \cem{\textbf{36.48↑}} \\
& Nemori~\cite{nan2025nemori} & 37.06 & 24.96 & 58.38 & 48.62 & 23.48 & 17.78 & 54.02 & 43.61 & 49.92 & 39.63 \\
& \cem{Nemori+\ours}* & \cem{\textbf{40.97}} & \cem{\textbf{29.36}} & \cem{\textbf{59.14}} & \cem{\textbf{50.01}} & \cem{\textbf{24.15}} & \cem{17.35} & \cem{\textbf{57.05}} & \cem{\textbf{46.46}} & \cem{\textbf{52.44↑}} & \cem{\textbf{42.25↑}} \\ \hline

\multirow{10}{*}{\textbf{GPT-4o}} 
& \textsc{LoCoMo~\cite{maharana2024evaluating}}$^{\dag}$ & 28.00 & 18.47 & 9.09 & 5.78 & 16.47 & 14.80 & \bs{61.56} & \bs{54.19} &41.67&35.10\\
& \textsc{ReadAgent~\cite{lee2024human}}$^{\dag}$& 14.61 & 9.95 & 4.16 & 3.19 & 8.84 & 8.37 & 12.46 & 10.29 &10.90 &8.63 \\
& \textsc{MemoryBank~\cite{zhong2024memorybank}}$^{\dag}$ & 6.49 & 4.69 & 2.47 & 2.43 & 6.43 & 5.30 & 8.28 & 7.10 &6.63 & 5.57\\
& \textsc{MemGPT~\cite{packer2023memgpt}}$^{\dag}$ & 30.36 & 22.83 & 17.29 & 13.18 & 12.24 & 11.87 & 60.16 & 53.35 & 42.78&36.80\\
\cline{2-12}
& A-MEM~\cite{xu2025mem}$^{\ddag}$ & 32.97 & 21.93 & 49.3 & 38.2 & 14.63 & 17.17 & 32.27 & 28.24 &  33.44 & 27.81 \\
& \cem{A-MEM+\ours}$^{\ddag}$* & \cem{28.03} & \cem{19.95} & \cem{\textbf{49.65}} & \cem{\textbf{38.61}} & \cem{14.17} & \cem{13.23} & \cem{\textbf{40.37}} & \cem{\textbf{35.63}} & \cem{\textbf{35.24↑}} & \cem{\textbf{29.88↑}} \\
& LightMEM~\cite{fang2025lightmem} &44.86 & 32.44 & 64.66 & 46.58 & 26.19 & 19.95 & 55.88 & 43.79  & 53.84 & 40.81 \\
& \cem{LightMEM+\ours}* & \cem{\textbf{46.35}} & \cem{\textbf{33.82}} & \cem{63.72} & \cem{45.95} & \cem{\textbf{28.08}} & \cem{\textbf{22.36}} & \cem{55.53} & \cem{43.53} & \cem{53.84} & \cem{\textbf{40.93↑}}

 \\
& Nemori~\cite{nan2025nemori} & 39.75 &28.92  & 55.86 & 47.30 & 23.22 & 17.28 & 56.48 & 46.17 & 51.21 &41.45 \\
& \cem{Nemori+\ours}* & \cem{39.52} & \cem{28.22} & \cem{\textbf{60.65}} & \cem{\textbf{51.21}} & \cem{\textbf{24.05}} & \cem{\textbf{18.10}} & \cem{56.39} & \cem{46.03} & \cem{\textbf{52.17↑}} & \cem{\textbf{42.11↑}} \\  
\hline
\end{tabular}%
}
\end{table*}

\noindent\textbf{Evaluation Metrics}. 
We employ F1 score, BLEU-1 score, and LLM-judge score to evaluate our approach. The F1 and BLEU-1 scores emphasize surface-level matching, including token-level overlap and n-gram overlap between the generated text and the ground truth, while the LLM score focuses on semantic correctness.


\noindent\textbf{Implementation Details}. All of our experiments are conducted on one RTX 4090D(24GB) GPU and one 15 vCPU Intel(R) Xeon(R) Platinum 8481C CPU. We select two representative LLMs as our backbones, including GPT-4o-mini and GPT-4o. Our method is implemented as a plug-in module and is applied to three recent and SOTA memory system approaches: A-MEM~\cite{xu2025mem}, LightMem~\cite{fang2025lightmem}, and Nemori~\cite{nan2025nemori}. A-MEM uses all-minilm-l6-v2 for text embedding. The pre-compressor model of LightMem is LLMLingua-2, and the embedding model of Nemori is text-embedding-3-small. We set the hyperparameters following the original version of the reports. Specifically, we conduct $k_a=10$ for the memory retrieval process of A-MEM. Meanwhile, the compression rate of LightMem is 0.6. For Nemori, the number of episodic memories retrieved in the searching step is 10, and that of semantic memories is 20. Our implementation of BLEU-1 computing is based on NLTK 3.9.2. For the three baselines A-MEM, LightMEM, and Nemori, we set the number of QA pairs generated per session to $k=10$, $k=3$, and $k=5$, respectively. The challenger, evaluator, and adapter share the same backbone as the baseline, using GPT-4o-mini or GPT-4o.

\begin{table*}[tb!]
\centering
\caption{
    Ablation Study on Nemori with GPT-4o-mini.
}
\label{tab:ablation}
\vspace{-5pt}

\setlength{\tabcolsep}{8pt} 
\renewcommand{\arraystretch}{1.2} 

\begin{tabular}{l| cc |cc| cc |cc| cc}
\hline
\multirow{2}{*}{\textbf{Model}}   
& \multicolumn{2}{c|}{\textbf{Multi-Hop}} 
& \multicolumn{2}{c|}{\textbf{Temporal}} 
& \multicolumn{2}{c|}{\textbf{Open-Domain}} 
& \multicolumn{2}{c|}{\textbf{Single-Hop}} 
& \multicolumn{2}{c}{\textbf{Average}} \\ 
\cline{2-11}
& \textbf{F1} & \textbf{BLEU} 
& \textbf{F1} & \textbf{BLEU} 
& \textbf{F1} & \textbf{BLEU} 
& \textbf{F1} & \textbf{BLEU} 
& \textbf{F1} & \textbf{BLEU} \\ 
\hline
\cem{Nemori+\ours} & \cem{\textbf{40.97}} & \cem{\textbf{29.36} }& \cem{59.14} & \cem{50.01} & \cem{24.15} & \cem{17.35} & \cem{\textbf{57.05}} & \cem{\textbf{46.46}} &\cem{\textbf{52.44↑}}& \cem{\textbf{42.25↑}} \\
 w/o Content Update& 36.86 &24.92 & 59.25 & 49.59 & \textbf{25.36} & \textbf{19.90} & 54.82 & 44.03 & 50.62 &40.19 \\ 
w/o Construction Update & 38.44 & 27.87 & 57.16&48.83 &20.89 & 16.00 & 45.72 & 44.79 &51.12 & 41.25 \\
    w/o Guided Question&39.81& 29.20& \textbf{59.77} & \textbf{50.69} & 21.16 & 16.19 & 54.58 & 44.26 & 50.87 & 41.09  \\
Nemori\cite{nan2025nemori} & 37.06 & 24.96 & 58.38 & 48.62 & 23.48 & 17.78 & 54.02 & 43.61 & 49.92 & 39.63 \\
\hline
\end{tabular}
\end{table*}

\begin{figure*}[t]
    \centering
    \resizebox{1\linewidth}{!}{
    \includegraphics[height=0.3\textwidth]{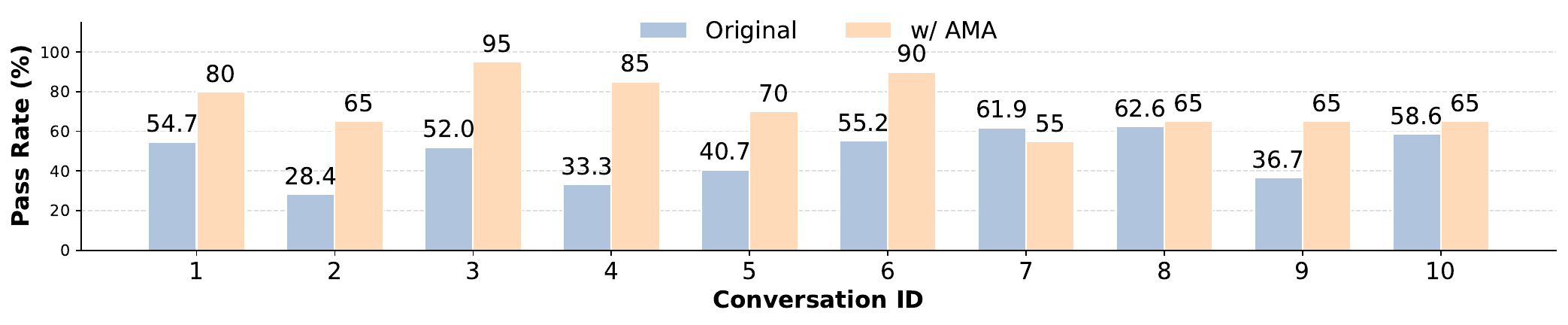}}
    \caption{The Evolutionary Process with \ours, with Nemori as the baseline and GPT-4o-mini as the backbone. 
    }
    \label{evolution}
    \vspace{-3mm}
\end{figure*}

\subsection{Main Results~(RQ1, 2)}
To evaluate how our method improves the memory quality and adaptability of existing memory systems~(\textbf{RQ1}), we conduct experiments on QA tasks across four categories from the LoCoMo dataset. To verify the compatibility of our mechanism~(\textbf{RQ2}), we integrate \ours into three baseline models and conduct experiments across two LLM backbones. Results in Table~\ref{tab:main} highlight the effectiveness, compatibility, and generalizability of \ours:
\begin{list}{$\bullet$}{
  \leftmargin=1em
  \itemsep=0pt
  \topsep=0pt
  \parsep=0pt
  \partopsep=0pt
}
\item \textbf{Overall Performance.} The method with \ours consistently outperforms state-of-the-art baselines in both F1 and BLEU metrics. Across most question categories and experimental settings, the integration of \ours leads to significant improvements, demonstrating its ability to enhance memory quality.
\item \textbf{Compatibility with Memory Systems.} The \ours mechanism is designed to flexibly integrate with existing memory systems, and it consistently improves performance when applied to the advanced memory systems A-MEM~\cite{xu2025mem}, LightMEM~\cite{fang2025lightmem}, and Nemori~\cite{nan2025nemori}. Specifically, for A-MEM, \ours brings notable gains, especially on GPT-4o-mini, where the average F1 score improves from 28.0 to 31.5, and BLEU from 24.4 to 27.3. For LightMEM, \ours enhances performance across most tasks, with average scores reaching 47.22 F1 and 36.48 BLEU on GPT-4o-mini. On Nemori, \ours further boosts strong baselines, achieving up to 52.44 F1 and 42.25 BLEU.  These results show that \ours can effectively strengthen various memory systems, despite the diversity of their underlying techniques.
\item \textbf{Generalizability across Backbone Models.} Beyond memory modules, \ours is also compatible with different backbone LLMs.
\ours achieves the best results on both GPT-4o-mini and GPT-4o. With Nemori, it reaches 52.44 F1 / 42.25 BLEU on GPT-4o-mini. With LightMEM, it reaches 53.84 F1 / 40.93 BLEU on GPT-4o, consistently outperforming all baseline methods.
The improvements observed across both backbones indicate that our method is not limited to a particular LLM backbone, and can be flexibly plugged into various foundation models. 
\item \textbf{Robustness Across Task Categories.} We observe that \ours brings steady gains across all four types of QA tasks.
In particular, tasks that require complex reasoning, such as Temporal and Open Domain, benefit significantly.
The most significant gains are observed in tasks involving more complex reasoning, such as Temporal and Open-Domain scenarios.
This suggests that our method is robust across diverse reasoning scenarios and can generalize to different QA tasks.
\end{list}

\begin{table*}[tb!]
\centering
\caption{
    Parameter Study on LightMEM using GPT-4o-mini.
}
\label{tab:parameter}
\vspace{-5pt}

\setlength{\tabcolsep}{9pt} 
\renewcommand{\arraystretch}{1.2} 

\begin{tabular}{l| cc |cc| cc |cc| cc}
\hline
\multirow{2}{*}{\textbf{Model}}   
& \multicolumn{2}{c|}{\textbf{Multi-Hop}} 
& \multicolumn{2}{c|}{\textbf{Temporal}} 
& \multicolumn{2}{c|}{\textbf{Open-Domain}} 
& \multicolumn{2}{c|}{\textbf{Single-Hop}} 
& \multicolumn{2}{c}{\textbf{Average}} \\ 
\cline{2-11}
& \textbf{F1} & \textbf{BLEU} 
& \textbf{F1} & \textbf{BLEU} 
& \textbf{F1} & \textbf{BLEU} 
& \textbf{F1} & \textbf{BLEU} 
& \textbf{F1} & \textbf{BLEU} \\ 
\hline
\cem{LightMEM+\ours (k=3)}& \cem{37.17} & \cem{\textbf{25.74}} & \cem{55.97} & \cem{\textbf{42.57}} & \cem{24.22} & \cem{18.17} & \cem{49.88} & \cem{39.84} & \cem{47.22↑} & \cem{\textbf{36.48}↑} \\
LightMEM+\ours (k=1)& 36.53 & 25.07 & 55.71 & 42.11 & 24.51 & 18.92 & 50.39 & 39.87 & 47.35↑ & 36.32↑ \\
LightMEM+\ours (k=10) & 36.10 & 25.72 & 56.82 & 42.32 & 23.72 & 17.6 & 50.27 & 39.53 & 47.39↑ & 36.22↑ \\
LightMEM~\cite{fang2025lightmem}&37.41 & 25.90 & 54.68 & 40.77 & 22.83 & 16.96 & 49.93 & 39.17 & 46.94 & 35.69\\
\hline
\end{tabular}
\end{table*}
\begin{figure*}[t]
    \centering
    \resizebox{1\linewidth}{!}{
    \includegraphics[height=0.3\textwidth]{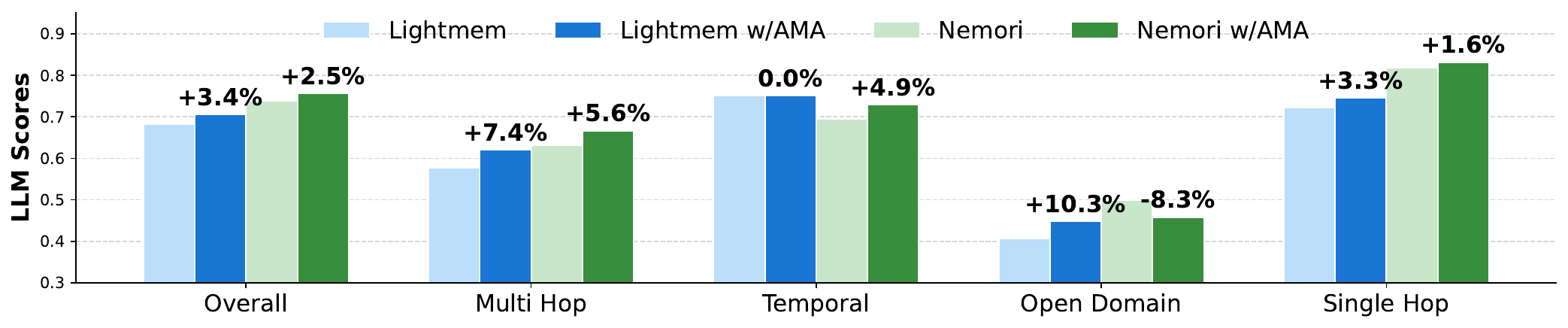}}
    \caption{LLM Evaluation Results on LightMEM and Nemori with GPT-4o-mini. 
    }
    \label{llm score}
    \vspace{-3mm}
\end{figure*}

\subsection{Ablation Study~(RQ3)}
To better investigate the contribution of each component in \ours (\textbf{RQ 3}), we conduct an ablation study with three controlled variants: \textbf{w/o content update}, \textbf{w/o construction update}, and \textbf{w/o guided question}. We observe that removing any single component leads to a performance drop in both F1 and BLEU scores across most task categories, demonstrating that each module contributes to the overall performance of the model.
The results are shown in Table~\ref{tab:ablation}.
\begin{list}{$\bullet$}{
  \leftmargin=1em
  \itemsep=0pt
  \topsep=0pt
  \parsep=0pt
  \partopsep=0pt
}
\item \textbf{w/o Content Update.} In this variant, we skip memory content update, which means the detected missing information is not integrated into the original memory. Without the content update mechanism, the system fails to recover missing information. As a result, the quality of generated answers declines, especially on the Multi-Hop and Single-Hop categories, where multi-turn reasoning or factual completeness is critical. 
\item \textbf{w/o Construction Update.} This variant omits the construction update mechanism and does not refine the memory extraction process. Without dynamic construction updates, the system lacks the ability to adapt its extraction process to different task scenarios. This makes the model fail to extract task-relevant information, leading to notable declines in performance, especially on Open-Domain questions where flexible reasoning is required. These results highlight the importance of integrating task preferences into the extraction process.
\item \textbf{w/o Guided Question.} In this variant, the Challenger is not guided by a structured prompt, which may lead to random and uncontrollable QA generation. Omitting guided question leads to the Challenger lacking explicit guidance on how to generate task-relevant questions.
The BLEU and F1 scores decrease slightly but still significantly outperform the baseline. This indicates that removing the guided question has negative impact on model performance. Nevertheless, our mechanism consistently improve the memory system, even when the generated questions do not exactly align with the task.
\end{list}

In summary, the ablation results confirm that each of the three components plays a distinct and essential role in enabling the memory system to generate high-quality memory. The removal of each component leads to a significant drop in performance, highlighting its individual contribution. Despite the performance degradation compared to the full \ours, all these variants still outperform the original baseline. This indicates that individual components of \ours consistently enhance the performance of the memory system. Furthermore, integrating all components yields the best overall results, demonstrating the importance of their joint interaction in enhancing memory and answer quality.

\subsection{The Evolutionary Process with AMA~(RQ4)}
To illustrate how the memory quality improves after applying \ours (\textbf{RQ4}), we conduct a follow-up test on each conversation. Specifically, we pose additional questions and handle them based on the updated memory, and compared the pass rates (i.e., answer accuracy) based on the original memory and the updated memory produced by our method (w/ AMA). As shown in the Figure~\ref{evolution}, the updated memory consistently shows higher pass rates across nearly all conversations. Specifically, in Conversation 2, the pass rate increases from 28.4\% to 65\%, and in Conversation 4, it rises from 33.3\% to 85\%. These improvements indicate that our memory update mechanism effectively enhances the completeness of memory, enabling the model to answer follow-up questions more accurately. Even in conversations where the original memory already had moderate performance (e.g., Conversation 8), the updated memory still provides slight gains. Overall, these results highlight that the quality of memory evolves through the dual-level update, leading to better downstream QA performance.

\subsection{Parameter Analysis~(RQ5)}
To investigate the effect of the number of QA pairs used during memory update~(\textbf{RQ5}), we conduct a parameter study by changing the number of QA pairs $k$ from the default value of 3 to 1 and 10. As shown in Table~\ref{tab:parameter}, our method remains robust to the number of QA pairs, achieving consistently strong performance.

Surprisingly, even when using only a single QA pair ($k=1$), the model achieves comparable results to the default setup. The average F1 and BLEU scores with $k=1$ (47.35 / 36.32) are very close to those with $k=3$ (47.22 / 36.48). This demonstrates that \ours is highly efficient, requiring only a small number of QA examples to effectively update memory.
Moreover, compared to the original baseline LightMEM, all three configurations ($k=1, 3, 10$) show noticeable improvements in both F1 and BLEU scores across all task types. This suggests that while the QA-driven update mechanism benefits from multiple examples, even minimal supervision is sufficient to guide meaningful memory evolution.

In summary, this parameter study confirms that \ours is not only effective but also resource-efficient, maintaining strong performance even under constrained QA signal conditions.
\subsection{LLM Evaluation Results~(RQ6)}
To further evaluate whether our method improves the quality of generated answers beyond traditional metrics~(\textbf{RQ6)}, we leverage GPT-4o-mini as evaluator to provide semantic-level assessments. Specifically, we compare \ours along with LightMEM and Nemori across four QA categories, using the LLM to score answer accuracy by assessing the semantic consistency between the original and generated answer.
As shown in the Figure~\ref{llm score}, baselines with \ours consistently achieve higher LLM scores across both the original methods. On average, the integration of our approach yields a +3.4\% improvement on LightMEM and a +2.5\% improvement on Nemori, confirming that our memory update mechanism leads to more semantically accurate and contextually appropriate answers.

In particular, in Category 1 (Multi Hop), which involves more complex or contextual reasoning, the gain is particularly large: +7.4\% on LightMEM and +5.6\% on Nemori. In Category 3, LightMEM with \ours significantly boosts its original LLM score by +10.3\%. 

These results demonstrate that our method helps improve the quality of memory, as validated by a strong LLM judge. It further confirms the effectiveness and generalizability of our approach in improving long-context QA performance.

\begin{table*}[t]
\centering
\caption{Case Study of Session-Level Memory Adaptation. For clarity, we omit information irrelevant to the questions from the raw dialogue and the memory (marked with ``...''), and highlight key information \textbf{in bold}.}
\begin{tabular}{p{2.6cm} | p{14.5cm}}
\toprule
\gre{\textbf{Raw Dialogue} }&
    \gre{Jon: Hey Gina! We haven't talked in a few days. \textbf{Been rehearsing hard and working on business plans.} It's been stressful, but dancing has kept me going.
...

Gina: Hah, yeah! \textbf{But really having a creative space for dancers is so important.} Last Friday at dance class with a group of friends I felt it. Your studio will be a go-to spot for self-expression. 
...

Gina: Remember Jon, \textbf{Just do it!}

Jon: Ah ha ha, yeah, JUST DOING IT! ...
 
}
 \\\hline

\yel{\textbf{Memory}

\textbf{Summary}} &
\yel{\textbf{Gina believes that a creative space for dancers is important} and felt that last Friday's dance class with a group of friends was a go-to spot for self-expression, encouraging the studio to keep up the good work and not forget the passion for dance. 
...
Jon committed to not quitting and to keep going, regardless of whatever comes his way.
\textbf{Gina reminded Jon to just do it.}
Jon responded affirmatively, stating that he is just doing it. ...}
 \\\hline
\cem{\textbf{QA 1}} &
\cem{Q:  What is Jon currently working on that has been stressful?

A~(G.T.): Jon has been rehearsing hard and working on business plans.

$\hat{A}$~(Memory): I cannot answer this question based on the available memory. \color{red} \textbf{$\times$}}
\\
\cem{\textbf{QA 2}} &
\cem{Q:  What phrase does Gina encourage Jon to remember?

A~(G.T.): Just do it!

$\hat{A}$~(Memory): Gina encourages Jon to remember to "just do it."
\color{red} \textbf{\checkmark}} \\
\cem{\textbf{QA 3}} &
\cem{Q:  What does Gina believe is important for dancers?

A~(G.T.): Having a creative space for dancers is important.

$\hat{A}$~(Memory): Gina believes that a creative space for dancers is important. \color{red} \textbf{\checkmark}}\\
\hline
\reddd{\textbf{Missing Summary}} &
\reddd{Jon is currently rehearsing hard and working on business plans.}\\
\hline

\reddd{\textbf{Improve }

\textbf{Instruction}}
& \reddd{Enhance memory extraction by focusing on detailed and contextual information, ensuring that all relevant aspects of a person's interests, motivations, and symbolic representations are captured.}
\\
\bottomrule
\end{tabular}
\label{tab:case-study0}
\vspace{-0.3cm}
\end{table*}

\subsection{Case Study}
To illustrate how our memory construction and update mechanism works, we present a case study that demonstrates the construction process of a session-level memory within the LightMEM system. 

From the raw dialogue, we generated three QA pairs targeting key pieces of information. These questions were then answered based on the constructed memory. As shown in Table~\ref{tab:case-study0}, the memory provides enough evidence to support Q2 and Q3, indicating that the system effectively captured Gina’s motivational phrase (``Just do it!'') and her belief in the importance of creative spaces for dancers.

However, for QA 1, which is about Jon’s current activities and stressors, the memory fails to provide an answer, despite the raw dialogue explicitly mentioning that Jon is ``rehearsing hard and working on business plans.'' This suggests that some important information was omitted during the memory construction.

To address this, we generated a missing summary that recovers the memory with the overlooked detail regarding Jon’s activities. Additionally, we propose an improvement instruction to guide subsequent construction processes.
This case highlights the importance of high-quality memory construction in supporting QA tasks and demonstrates how \ours can identify issues and iteratively improve through targeted memory updates and guided refinement.






\section{Related Work}
In this section, to illustrate how offline memory preparation is performed in existing memory systems and to facilitate a better understanding of our adaptation mechanism, we review related techniques for memory construction and memory update.
\subsection{Memory Construction}
Memory construction aims to extract key information from long-term dialogue and organize it into structured representations and is a fundamental component of the memory system~\cite{chhikara2025mem0buildingproductionreadyai,park2023generative}.
High-quality memory helps dialogue systems maintain contextual consistency and directly affects subsequent retrieval quality~\cite{sun2025hierar,wu2025longmemeval,hong2024metagpt}.

Existing methods typically employ techniques such as summarization and keyword/entity extraction to obtain compressed representations of information from the original dialogue~\cite{kang2025memoryosaiagent,jiang2025memory}.
For example, Rsum~\cite{wang2025recursively} recursively generates summaries, starting with short dialogues and then integrating the existing summaries with subsequent conversations to produce high-level summaries.
 A-MEM~\cite{xu2025mem} extracts keywords from the dialogue to capture key concepts. In terms of memory organization, different methods adopt structures with varying levels of granularity and hierarchy. SECOM~\cite{pan2025memory} introduces a dialogue segmentation approach, constructing memory using segments as the smallest unit to capture logical relationships among units. LightMEM~\cite{fang2025lightmem} clusters and restructures dialogue content based on topics, forming a topic-aware short-term memory module that enhances the modeling of current dialogue intentions.
In terms of memory storage, the memory entries can be encoded as natural language text or structured graphs to enhance retrieval capabilities. ZEP~\cite{rasmussen2025zep} employs Temporal Knowledge Graphs (TKG) to capture the evolution of knowledge throughout the dialogue, thereby supporting time-sensitive information retrieval. AssoMem~\cite{zhang2025assomemscalablememoryqa} constructs an Associative Memory Graph, where dialogue segments are linked according to diverse attributes such as speakers, intents, and motivations, forming a graph-structured memory that facilitates reasoning.

Although the aforementioned methods have made significant progress in information extraction, memory organization, and memory storage, they typically rely on predefined, manually designed pipelines such as fixed extraction rules, summarization strategies, or structural templates. This rigid design limits the flexibility and adaptability of memory construction in handling diverse tasks and domain-specific requirements. To mitigate this limitation, our \ours adapts extraction strategies to capture task-specific information, thereby providing sufficient evidence for downstream reasoning. 

\subsection{Memory Update and Management}
Memory update refers to the process of modifying existing memory content during the construction of new memory from ongoing dialogue, in order to resolve memory conflicts or knowledge obsolescence~\cite{wang2024wise,wang2025mextending,shen2024encode,wang2025mem}. 
Existing approaches often perform conflict detection, generate new links, or simulate human memory mechanisms such as forgetting and consolidation to maintain memory relevance and consistency. For example, MemoryBank~\cite{zhong2024memorybank} is inspired by Ebbinghaus' forgetting theory and updates memory by selectively removing or reinforcing content based on time intervals or memory priority. 
A-MEM~\cite{xu2025mem} executes updates to existing memory based on new dialogue and establishes new links among memory entries. It automatically retrieves relevant historical memories and establishes semantic associations, while incorporating a memory evolution mechanism to form an interconnected and continuously evolving memory architecture for intelligent agents. 
However, these update mechanisms are driven by indirect indicators, rather than being directly evaluated based on task-specific objectives. As a result, the memory fails to align with the task demands, which may negatively impact downstream performance. Our method enhances memory quality by explicitly updating missing task-specific information.
\section{Conclusion}
While existing memory systems have achieved significant progress, there remains a misalignment between their construction/update phases and specific task demands. To address this issue, we propose an \textbf{A}dversarial \textbf{M}emory \textbf{A}daptation~(AMA) mechanism that simulates task supervision and adaptively updates both memory construction strategy and content based on feedback. Our mechanism can be seamlessly integrated into existing memory systems. Extensive experiments on long-dialogue benchmark LoCoMo validate the effectiveness of our approach.
\bibliographystyle{ACM-Reference-Format}
\bibliography{main}

\end{document}